\documentclass[letterpaper, 10 pt, conference]{ieeeconf}  

\IEEEoverridecommandlockouts                              

\usepackage{lmodern}
\usepackage{graphicx}
\usepackage{times}
\usepackage[caption=false]{subfig}
\usepackage{multicol}
\usepackage{multirow}
\usepackage[font=small]{caption}
\usepackage{amsfonts}
\usepackage{amsmath}
\usepackage{algpseudocode}
\usepackage{algorithm}
\usepackage{listings}
\usepackage{cite}
\usepackage{siunitx,booktabs}
\usepackage{comment}

\newcommand{\etal}{\textit{et al}.}

\title{\LARGE \bf
Adversarial Attack and Defense of YOLO Detectors\\ in Autonomous Driving Scenarios
}

\author{Jung Im Choi$^{1}$ and Qing Tian$^{2}$%
\thanks{$^{1}$Jung Im Choi is a PhD candidate in Data Science, Bowling Green State University, Bowling Green, OH 43403, USA.
{\tt\small choij@bgsu.edu}}%
\thanks{$^{2}$Qing Tian (corresponding author) is an assistant professor at the Department of Computer Science, Bowling Green State University, Bowling Green, OH 43403, USA. {\tt\small qtian@bgsu.edu}}%
}

\begin{document}

\bibliographystyle{ieeetr}

\maketitle
\thispagestyle{empty}
\pagestyle{empty}

\begin{abstract}

Visual detection is a key task in autonomous driving, and it serves as a crucial foundation for self-driving planning and control. Deep neural networks have achieved promising results in various visual tasks, but they are known to be vulnerable to adversarial attacks. A comprehensive understanding of deep visual detectors' vulnerability is required before people can improve their robustness. However, only a few adversarial attack/defense works have focused on object detection, and most of them employed only classification and/or localization losses, ignoring the objectness aspect. In this paper, we identify a serious objectness-related adversarial vulnerability in YOLO detectors and present an effective attack strategy targeting the objectness aspect of visual detection in autonomous vehicles. Furthermore, to address such vulnerability, we propose a new objectness-aware adversarial training approach for visual detection. Experiments show that the proposed attack targeting the objectness aspect is 45.17\% and 43.50\% more effective than those generated from classification and/or localization losses on the KITTI and COCO\_traffic datasets, respectively. Also, the proposed adversarial defense approach can improve the detectors' robustness against objectness-oriented attacks by up to 21\% and 12\% mAP on KITTI and COCO\_traffic, respectively.

\end{abstract}

\section{INTRODUCTION}

Over the past decade, deep learning has revolutionized various visual computing areas, such as object detection~\cite{Ren2015,Redmon2016}, image classification~\cite{Szegedy2014,Simonyan2014}, image captioning~\cite{You2016}. Vision-based self-driving cars can take advantage of deep neural networks to better detect objects (e.g., cars, road signs, etc.)~\cite{Fujiyoshi2019,Grigorescu2020}. However, deep learning models can easily fall victim to adversarial attacks~\cite{Goodfellow2015,Carlini2016,Xie2017,Lu2017}. While numerous adversarial robustness studies have targeted classification models~\cite{Metzen2017,Liao2017,Samangouei2018}, few have focused on the more challenging task of object detection, especially in autonomous driving scenarios.

Unlike image classification which only requires one class label for an entire image, object detection involves three types of outputs for each region of interest in an input image: (1) the objectness (the probability of the associated bounding box containing an object), (2) the bounding box location, and (3) the class label. Due to the high complexity, object detectors can be more difficult to attack and defend compared to classification. Thus, a deeper and more holistic understanding of object detectors' vulnerability is needed before we can improve their robustness.
Some research has been carried out targeting two-stage detectors~\cite{Girshick2013,Ren2015}. Most of them (e.g., \cite{Xie2017, Chen2018}) consider only the tasks in the second stage (i.e., localization and classification), with the objectness aspect in the first stage ignored.
Compared to the attack works on two-stage detectors, not many efforts have been made to investigate one-stage detectors (e.g., YOLOs~\cite{Redmon2016}), which is more suitable for autonomous driving scenarios due to its high speed and efficiency.

\begin{figure}
    \centering
    \includegraphics[width=.9\linewidth, height=2.1in]{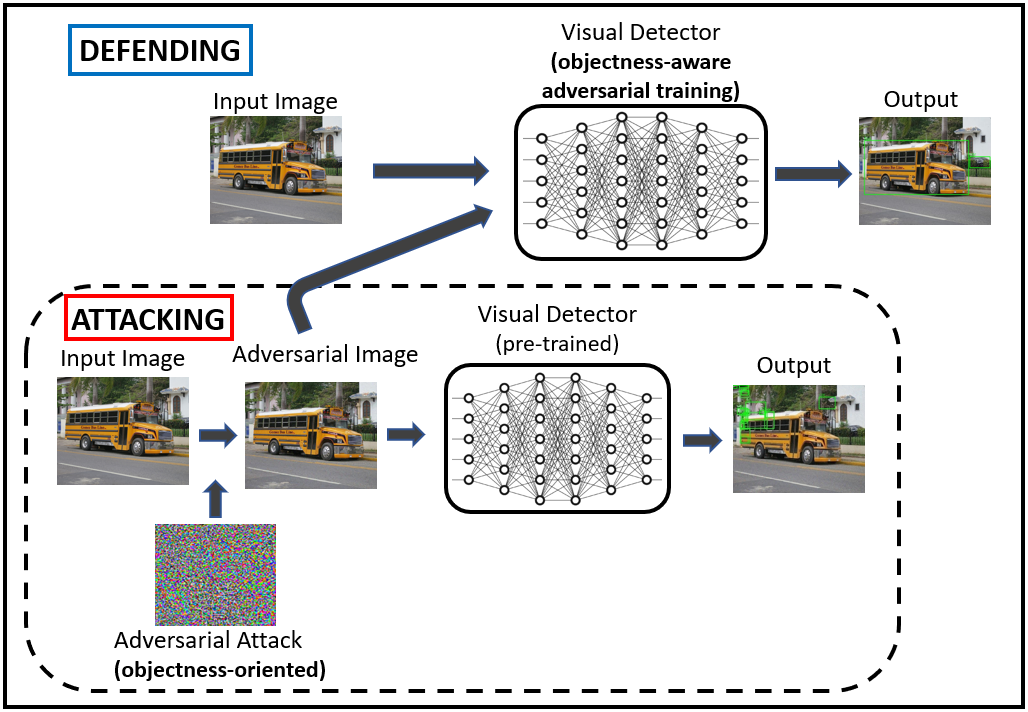}
    \caption{A schematic overview of the proposed object-oriented attacking and
defending strategies. Our objectness-oriented attacking and defending approaches are more effective than the existing methods that utilize only classification and/or localization losses.}
    \label{fig:overview}
\end{figure}

In this paper, we propose a more effective attack strategy that takes into account the objectness aspect of object detection in self-driving cars. Our approach is designed for the state-of-the-art YOLO detector (i.e., YOLOv4~\cite{Bochkovskiy2020}), although it is likely to be applicable to other detectors as well. To defend the designed attack, we also propose a new objectness-aware adversarial training strategy. Figure~\ref{fig:overview} provides a schematic overview of our proposed attacking and defending approaches. Our experiments on the KITTI~\cite{Geiger2012} and COCO\_traffic (a subset of COCO~\cite{Lin2014}) datasets demonstrate that attacks based on the objectness loss are more effective than those based on other task losses for object detection in self-driving scenarios. In summary, the main contributions of this paper are as follows:
\begin{itemize}
    \item We identify a serious adversarial vulnerability in YOLO detectors by evaluating the impact of adversarial attacks sourced from the multiple task losses (i.e., of objectness, localization, and classification) and present an effective attack approach targeting the objectness aspect of YOLO in autonomous driving. The objectness-oriented attacks can be more effective than those generated from classification and/or localization losses.
    \item Based on our analysis of objectness-related vulnerability, we also propose a new adversarial training-based strategy utilizing the objectness loss. The model trained with objectness-based attacks can be more robust than those utilizing other two task losses against the task-oriented attacks.
    \item Our objectness-aware adversarial training can help alleviate the potential conflicts/misalignment of the directions of the image gradients derived from different task losses in object detection.
\end{itemize}

\section{RELATED WORK} \label{relatedWork}

\subsection{Adversarial Attacks for Visual Detection}
Several studies have shown that slightly perturbing an original image can fool a target model to produce wrong predictions~\cite{Goodfellow2015,Carlini2016}. While most of the existing adversarial attacks are designed for classification models~\cite{Metzen2017,Liao2017,Samangouei2018}, relatively few works have focused on the more challenging object detection task~\cite{Xie2017,Lu2017}. Depending on whether the attacker has access to the victim model's internal detail (e.g., parameters), adversarial attacks can be categorized into white-box~\cite{Goodfellow2015,Carlini2016,Xie2017} or black-box~\cite{Papernot2016,Guo2019} attacks. In this paper, we will consider white-box attacks. Xie \etal~\cite{Xie2017} extended the attack generation method from classification to object detection by using the dense adversary generation. Lu \etal~\cite{Lu2017} created adversarial examples for detectors by generating many proposals and randomly assigning a label for each proposal region. However, \cite{Xie2017} and \cite{Lu2017} used only the classification loss to generate their adversarial examples for given target detectors. Li \etal~\cite{Li2018} presented a robust adversarial perturbation method to attack the Region Proposal Network (RPN) by incorporating both classification and localization losses. Zhang \etal~\cite{Zhang2019} identified an asymmetric role of classification and localization losses and find that the image gradients from the two losses are misaligned, which can make effective adversary generation difficult. Unlike the existing attack methods which used classification and/or localization loss, we propose to leverage objectness loss to generate effective adversarial examples for visual detection in self-driving scenarios which will be shown to be more effective.

\subsection{Adversarial Training for Visual Detectors}
Adversarial training~\cite{Goodfellow2015, Madry2019} is one of the most effective approaches to defend deep learning models against adversarial attacks~\cite{Athalye2018}. Since its introduction by Szegedy \etal~\cite{Szegedy2014}, many effective defense methods~\cite{Madry2019,Tramer2020} for classification have been proposed. While those approaches greatly increased the adversarial robustness of deep classifiers, not many efforts have been made to improve the robustness of object detectors, especially in safety-critical autonomous driving scenarios. Recently, Zhang \etal~\cite{Zhang2019} have generalized the adversarial training framework from classification to object detection. They utilized the task-oriented domain constraint in adversarial training to improve the robustness of object detectors. Chen \etal~\cite{Chen2021} presented Det-AdvProp which employs separate batch normalization layer for clean images and adversarial examples to address the mAP score decrease of adversarially trained model on clean images. However, both of the above-mentioned approaches considered only localization and classification losses. In this paper, we will analyze and leverage all of the three task losses in the YOLOv4 detector (i.e., objectness, localization, and classification losses) during adversarial training.
By explicitly considering the objectness aspect, our adversarial training method can better align the image gradients sourced from different objective components and thus lead to more robust visual detectors.

\section{METHODOLOGY} \label{sec:method}

\subsection{Adversarial Vulnerability in YOLO Detectors}
In this subsection, we examine various aspects of the YOLO detector for potential adversarial susceptibility and identify a serious vulnerability in the objectness component. We take YOLOv4~\cite{Bochkovskiy2020}, a state-of-the-art variant in the YOLO family, as the base model. Compared to two-stage detectors, it is more efficient and suitable for vision-based self-driving systems. While two-stage detectors propose regions of interest (ROI) before classifying and regressing bounding boxes, YOLOv4 tackles classification and regression in a single stage without any ROI proposal step. In YOLOv4, the overall loss for each box prediction consists of three components, i.e., objectness, localization, and classification losses:
\begin{equation} \label{eq:1}
\resizebox{.975\hsize}{!}{$\mathcal{L}(\textbf{x},y,\textbf{b};\theta) = \mathcal{L}_{OBJ}(\textbf{x},\textbf{b};\theta)+\mathcal{L}_{LOC}(\textbf{x},\textbf{b};\theta)+\mathcal{L}_{CLS}(\textbf{x},y;\theta)$},
\vspace{0.1in}
\end{equation}

\noindent where $\textbf{x}$ is the training sample, $y$ and $\textbf{b}$ are the ground-truth class label and bounding box, $\theta$ represents the model parameters, and $\mathcal{L}(\cdot)$ indicates a loss
function. The subscripts stand for the three aspects in object detection (e.g., $OBJ$ for objectness, $LOC$ for localization, and $CLS$ for classification). While most existing works exploring adversarial robustness (e.g., \cite{Chen2018, Zhang2019, Chen2021}) have been focused on utilizing the classification and/or localization losses, we argue that effective attacks for object detection should consider all of the three aspects, including the usually ignored objectness loss. The objectness loss, which is the main focus of the paper, can be divided into two parts: the object (obj) part and the no\_object (no\_obj) part:

\begin{equation} \label{eq:objectnessloss}
\mathcal{L}_{OBJ}(\textbf{x},\textbf{b};\theta)=\mathcal{L}_{obj}(\textbf{x},\textbf{b};\theta)+\lambda_{no\_obj}\,\mathcal{L}_{no\_obj}(\textbf{x},\textbf{b};\theta),
\end{equation}
\noindent where
\begin{equation}
    \begin{split}
    \mathcal{L}_{obj}(\textbf{x},\textbf{b};\theta) = & -\sum_{k=1}^{K}I_{k}^{obj} \bigg[ \hat{C}_k \log(C_k)\\
    &+(1-\hat{C}_k) \log(1-C_k) \bigg],
    \end{split}
\end{equation}
\noindent and
\begin{equation}
    \begin{split}
    \mathcal{L}_{no\_obj}(\textbf{x},\textbf{b};\theta)=& -\sum_{k=1}^{K}I_{k}^{no\_obj} \bigg[ \hat{C}_k \log(C_k)\\
    &+(1-\hat{C}_k) \log(1-C_k) \bigg].
    \end{split}
\end{equation}

\noindent The objectness score $\hat{C}_k \in [0,1]$ can be considered as the network’s confidence in a given bounding box containing an object. 
$\lambda_{no\_obj} \in [0,1]$ is a hyperparameter penalizing no-object bounding boxes (according to the ground truth), $K$ is the number of predicted bounding boxes, and $I_{k}^{obj}$ represents whether the $k$-th bounding box contains an object (i.e., $I_{k}^{obj}$ = 1) or not (i.e., $I_{k}^{obj}$ = 0). Similarly, $I_{k}^{no\_obj}$ = 1 denotes the $k$-th bounding box has no object.

The localization loss ${L}_{CLS}$, responsible for finding the bounding-box coordinates, is based on the Complete Intersection over Union (CIoU) loss~\cite{Zheng2020}:
\begin{equation} \label{eq:3}
\mathcal{L}_{LOC}(\textbf{x},\textbf{b};\theta)=\mathcal{L}_{CIoU}=1-IoU+\frac{\rho^2(\hat{\textbf{b}},\textbf{b})}{c^2}+\alpha v,
\end{equation}
where IoU (Intersection over Union) is an evaluation metric used to measure overlap between two bounding boxes, $\rho(\hat{\textbf{b}},\textbf{b})$ represents the Euclidean distance of central points of the prediction box $\hat{\textbf{b}}$ and the ground truth $\textbf{b}$, $c$ is the diagonal distance of the smallest enclosing box covering $\hat{\textbf{b}}$ and $\textbf{b}$, $\alpha$ is a positive trade-off hyperparameter, and $v$ is the consistency measure of aspect ratio.

The classification loss ${L}_{CLS}$, responsible for the class-score prediction $\hat{\textbf{p}}_k(i)$, is defined as:
\begin{equation} \label{eq:4}
\begin{split}
\mathcal{L}_{CLS}(\textbf{x},y;\theta) &= -\sum_{k=1}^{K}I_{k}^{obj}\sum_{i\in classes}\bigg[\hat{p}_k(i) \log(p_k(i))\\
&\quad + (1-\hat{p}_k(i)) \log(1-p_k(i)) \bigg].
\end{split}
\end{equation}

\begin{figure}[t!]
    \centering
    \includegraphics[width=.9\linewidth, height=1.8in]{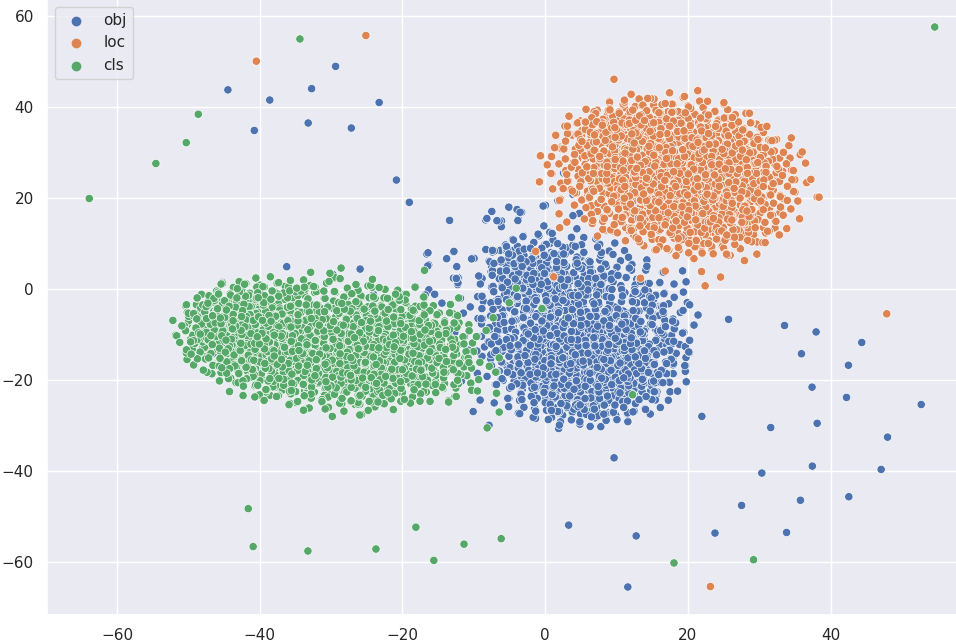}
    \caption{Visualization of adversarial examples generated from different task losses using t-SNE. Different colors encode the task losses used for generating adversarial examples (blue: objectness, orange: localization, green: classification loss). This is a typical example from the KITTI dataset for autonomous driving.}
    \label{fig:kiiti t-SNE}
\end{figure}

To see how adversarial samples derived from the different losses are distributed, we project their high-dimensional representations into a 2D space by t-SNE and show the task gradient domains in Figure~\ref{fig:kiiti t-SNE}. 
Given a clean image, each dot in the figure represents one adversarial example derived from one of the three task losses (i.e., $\mathcal{L}_{OBJ}$, $\mathcal{L}_{LOC}$, and $\mathcal{L}_{CLS}$). Interestingly, we observe that the objectness-based gradient domain (shown in blue) partially overlaps with both the classification-based (shown in green) and localization-based (shown in orange) gradient domains while there is no overlapping between the classification and localization domains\footnote{From a probabilistic point of view, no overlapping between the two does not mean that no attack can simultaneously handle the two aspects. However, the chance is low (it may be possible with more samples).}. Non-overlapping regions of the task gradient domains reflect the inconsistent directions of the image gradients derived from the task losses (we will refer to this issue as `misaligned task gradients' for the rest of the paper). It is worth mentioning that Zhang \etal~\cite{Zhang2019} found similar issues in general object detection considering only the classification and localization domains. They attempted to avoid such conflicts/misalignment by choosing one of the two types of attacks (either classification or localization oriented) each time. However, an attack from one domain is likely to ignore the other aspect, especially in our autonomous driving scenarios where the two regions (orange and green) are far apart (e.g., Figure~\ref{fig:kiiti t-SNE}). The chance is low that an adversarial example derived from the classification or localization loss can simultaneously attack both aspects.
The objectness domain, lying between the classification and localization domains, helps join the other two aspects and attract more attention to the middle regions where an attack has a better chance to target all three aspects. 
Figure~\ref{fig:kiiti t-SNE} visually demonstrates the objectness-related vulnerability in the YOLO detector and inspires us to employ the objectness loss to generate more effective adversarial attacks for object detection.  

\subsection{Objectness-Oriented Adversarial Attack for Visual Detection}
Motivated by the preceding analysis, we propose to consider all the three loss/vulnerability aspects and utilize the objectness loss to craft adversarial attacks. Given a trained deep learning model $f$ and an input $\textbf{x}$, generating an adversarial example $\textbf{x}^{\prime}$ can be formulated as:
\begin{equation} \label{eq:5}
||\textbf{x}^{\prime}-\textbf{x}||_p<\epsilon  \quad s.t.\quad f(\textbf{x}^{\prime})\neq f(\textbf{x}),
\end{equation}
where $||\cdot||_p$ denotes the distance ($L_p$ norm) between two data sample. The choice of the norm, $p$, determines the type of limitations placed on the adversary generation. As in many previous works (e.g.,~\cite{Zhang2019}~\cite{Chen2021}), we utilize $L_{\infty}$ as a distortion measure, and it measures the maximum absolute change to any pixel. The attack budget $\epsilon$ bounds the maximum perturbation in terms of $L_{\infty}$. Through exploring in the original data space, this optimization process tries to produce an incorrect prediction while being subject to a constraint on the perturbation magnitude. To generate adversarial examples, we take gradients of the corresponding losses (i.e., objectness, localization, and classification losses) with respect to the input and modify the input along the gradient direction:
\begin{equation} \label{eq:6}
\begin{gathered}[b]
\textbf{x}^{\prime}_{obj,PGD}=\mathcal{P}(\textbf{x}+\alpha\cdot sign(\bigtriangledown_{\textbf{x}}\mathcal{L}_{OBJ}(\textbf{x},\textbf{b};\theta))),\\
\textbf{x}^{\prime}_{loc,PGD}=\mathcal{P}(\textbf{x}+\alpha\cdot sign(\bigtriangledown_{\textbf{x}}\mathcal{L}_{LOC}(\textbf{x},\textbf{b};\theta))),\\
\textbf{x}^{\prime}_{cls,PGD}=\mathcal{P}(\textbf{x}+\alpha\cdot sign(\bigtriangledown_{\textbf{x}}\mathcal{L}_{CLS}(\textbf{x},y;\theta))),
\end{gathered}
\end{equation}
where $\mathcal{P}$ projects the perturbed example to a $\epsilon$-radius ball $\{\textbf{x}\, |\, ||\textbf{x}^{\prime}-\textbf{x}||_{\infty} \leq \epsilon\}$ to ensure the perceptual similarity, and $\alpha$ represents the step size.
Note that if the number of iterations in PGD~\cite{Madry2019} equals to one, it becomes the FGSM method~\cite{Goodfellow2015}. We will explore both in our experiments. 

\subsection{Objectness-Aware Adversarial Training}
On the defense side, to improve the adversarial robustness of object detectors, we develop a new objectness-aware adversarial training approach explicitly utilizing the objectness aspect mentioned in the previous subsections. Unlike prior works~\cite{Zhang2019,Chen2021} where models were trained with attacks generated from the localization and/or classification losses only, our approach considers all dimensions of the object detection output and uses all the heterogeneous sources of losses (i.e., objectness, localization, and classification tasks) in the adversary generation and adversarial training. The overall objective of the proposed adversarial training can be defined as follows:    
\begin{equation} \label{eq:8}
\mathop{\textrm{arg min}}_{\theta}\mathbb{E}_{\textbf{x}\sim \mathcal{D};y,\textbf{b}\sim \mathcal{B}\,(\textbf{x})}\mathcal{L}(\textbf{x},\{y,\textbf{b}\};\theta)+\mathcal{L}(\underline{\textbf{x}},\{y,\textbf{b}\};\theta),\\
\end{equation}
where $\underline{\textbf{x}}$ is the strongest one among the adversarial examples generated from the three task losses in terms of the overall loss. Algorithm~\ref{alg:Adv.Train} shows the details of the proposed adversarial training algorithm. 
\begin{algorithm}
	\caption{Objectness-Aware Adversarial Training} 
	\label{alg:Adv.Train} 
	\textbf{Input:} Dataset $\mathcal{D}$, Training epochs $N$, Batch size $B$, Perturbation bounds $\epsilon$ \begin{algorithmic}
	\For{epoch = 1 \textbf{to} $N$}
	    \For{random batch \{$\textbf{x}^i,\{y^i,\textbf{b}^i\}\}_{i=1}^B\sim \mathcal{D}$}
	        \State  
	        $(\textbf{x}^{\prime}_{obj})^i=\mathcal{P}(\textbf{x}^i+\epsilon\cdot sign(\bigtriangledown_{\textbf{x}}\mathcal{L}_{OBJ}(\textbf{x}^i,\textbf{b}^i;\theta)))$        
	        \State  $(\textbf{x}^{\prime}_{loc})^i=\mathcal{P}(\textbf{x}^i+\epsilon\cdot sign(\bigtriangledown_{\textbf{x}}\mathcal{L}_{LOC}(\textbf{x}^i,\textbf{b}^i;\theta)))$
	        \State  $(\textbf{x}^{\prime}_{cls})^i=\mathcal{P}(\textbf{x}^i+\epsilon\cdot sign(\bigtriangledown_{\textbf{x}}\mathcal{L}_{CLS}(\textbf{x}^i,y^i;\theta)))$
	        \State Choose $\underline{\textbf{x}}^i$ that leads to the max total loss:
	        \State $\underline{\textbf{x}}^i=\underset{\tilde{\textbf{x}}^i\in \{(\textbf{x}^{\prime}_{obj})^i,(\textbf{x}^{\prime}_{loc})^i,(\textbf{x}^{\prime}_{cls})^i\}}{\textrm{arg max}}\mathcal{L}(\tilde{\textbf{x}}^i,\{y^i,\textbf{b}^i\};\theta)$
	         \State Perform an adversarial training step w.r.t. $\theta$:
	        \State $\mathop{\textrm{arg min}}_{\theta}$  $\mathcal{L}(\textbf{x}^i,\{y^i,\textbf{b}^i\};\theta)+\mathcal{L}(\underline{\textbf{x}}^i,\{y^i,\textbf{b}^i\};\theta)$
	   \EndFor
	\EndFor
    \end{algorithmic}
    \textbf{Output:} Learned model parameter $\theta$
\end{algorithm}

We first search for the most detrimental adversarial perturbation from the three candidates. Then, we update the model parameters to reduce the overall loss on both a clean example and the selected adversarial example $\underline{\textbf{x}}$. We use a similar max-max scheme to Zhang \etal~\cite{Zhang2019} and keep the adversarial example (out of three) that maximizes the overall loss. However, our approach is different from their work in that we include the critical objectness component. In their work, each time, only one type of attack is chosen (e.g., either classification or localization). As shown in Figure~\ref{fig:kiiti t-SNE}, there can be hardly any overlapping between the two task domains in autonomous driving scenarios that we care about. It follows that choosing one type of attack likely means ignoring the other vulnerability aspect. Improved adversarial robustness towards one task domain does not necessarily reflect the overall model robustness (the adversarial robustness towards the other aspect may be reduced). Smart attackers may attack both aspects simultaneously, and we need a more comprehensive defense strategy.
Including the objectness component helps fill in the missing piece. Utilizing the adversarial example derived from the objectness loss can better alleviate the issue of misaligned task gradients (Figure~\ref{fig:kiiti t-SNE}). Its generated attack can potentially be more detrimental and adversarial training taking into consideration such examples is more helpful to improve the model robustness. 
We adapt FGSM-based adversarial training combined with random initialization~\cite{Wong2020}, which is as effective as PGD-based training but has significantly lower computational cost.  
Experimental results will demonstrate our strategy's efficacy in the following section (Sec.~\ref{sec:experiments}).

\section{EXPERIMENTS AND RESULTS} \label{sec:experiments}

\subsection{Experimental Setup}
In our experiments, we use the one-stage object detector, YOLOv4~\cite{Bochkovskiy2020} as the base model. For its backbone feature extraction network, CSPDarkNet53 is used, which is pretrained on COCO2017~\cite{Lin2014} dataset. We conduct our experiments on both the KITTI~\cite{Geiger2012} and COCO\_traffic datasets. The latter is a subset of MS-COCO~\cite{Lin2014}. For the KITTI dataset, we follow the same convention for combining the categories and splitting the dataset as in~\cite{Wu2017}. To be more specific, there are three categories for the KITTI dataset: car, cyclist, and pedestrian. The 7,481 training images are split in half into a training set and a validation set since the test images do not have labels. For the COCO\_traffic dataset, it has 8 categories related to autonomous driving (i.e., person, bicycle, car, motorcycle, bus, truck, traffic light and stop sign). There are 71,536 training images and 3,028 test images. For both datasets, the YOLOv4 is trained to convergence (50 epochs) using an Adam optimizer, with an initial learning rate of 0.001 and a batch size of 8. Then, we evaluate the models using the Pascal VOC mean average precision (mAP) metric with the IoU threshold set as 0.5. 

Fast Gradient Sign Method (FGSM)~\cite{Goodfellow2015} and Projected Gradient Descent (PGD)~\cite{Madry2019} are two well-known attacking methods originally designed for classification tasks. We extend them to object detection scenarios through combining them with the losses in Equation~\ref{eq:6}. For each type of attack, a range of attacking strengths $\epsilon \in \{2,4,6,8\}$ are considered. For PGD, we use a step size $\alpha=1$ and the number of iterations $T=10$. For adversarial training, we adapt the FGSM algorithm (with $\epsilon=4$). It generates the adversarial examples which will be used as input to adversarial training.

\subsection{Quantitative Analysis of Attacks}
In this subsection, we investigate model vulnerability to a variety of task-oriented attacks. Table~\ref{tab:kitti att. size} and Table~\ref{tab:coco att. size} demonstrate the mAP changes on the two datasets due to the attacks of different sources ($\mathcal{A}_{loc}, \mathcal{A}_{cls},\mathcal{A}_{obj}$), types (FGSM, PGD) and strengths ($\epsilon \in \{2,4,6,8\}$). In our experiments, the attack strengths are determined by different maximum perturbations, where larger perturbation budget indicates stronger attack.

\begin{table}[h!]
\centering
\setlength\tabcolsep{5pt}
\begin{tabular}{@{} l*{5}{S} @{}}
\toprule
{Method} & {Att.Size} & {$\mathcal{A}_{loc}$} & {$\mathcal{A}_{cls}$} & {$\mathcal{A}_{obj+loc+cls}$} & {$\mathcal{A}_{obj}$}\\
\midrule
\multirow{4}{*}{FGSM} & {$\epsilon=2$} 
& -0.98 & -0.97 & -8.42 & \textbf{-10.49}\\
& {$\epsilon=4$} & -3.20 & -3.15 & -13.88 & \textbf{-16.85} \\
& {$\epsilon=6$} & -6.08 & -5.65 & -17.88 & \textbf{-22.53} \\
& {$\epsilon=8$} & -10.44 & -9.65 & -22.04 & \textbf{-27.31} \\
\midrule
\multirow{4}{*}{PGD-10} & {$\epsilon=2$} & -1.22 & -0.87 & -42.44 & \textbf{-42.64} \\
& {$\epsilon=4$} & -4.11 & -2.64 & -51.47 & \textbf{-51.67} \\
& {$\epsilon=6$} & -7.00 & -5.91 & -54.17 & \textbf{-54.39} \\ 
& {$\epsilon=8$} & -10.66 & -9.59 & -55.48 & \textbf{-55.83} \\
\bottomrule
\end{tabular}
\caption{Model performance degradation under various attack strengths for the task-oriented attacks using FGSM and 10-step PGD on KITTI. $\mathcal{A}_{loc}, \mathcal{A}_{cls}, \mathcal{A}_{obj+loc+cls}, and~\mathcal{A}_{obj}$ denote the attacks sourced from corresponding task losses (i.e., localization, classification, overall, and objectness losses). The objectness-oriented attacks decrease the mAP most. The clean mAP on KITTI is 80.10\%.}
\label{tab:kitti att. size}
\end{table}

\begin{table}
\centering
\setlength\tabcolsep{5pt}
\begin{tabular}{@{} l*{5}{S} @{}}
\toprule
{Method} & {Att.Size} & {$\mathcal{A}_{loc}$} & {$\mathcal{A}_{cls}$} & {$\mathcal{A}_{obj+loc+cls}$} & {$\mathcal{A}_{obj}$}\\
\midrule
\multirow{4}{*}{FGSM} & {$\epsilon=2$} & -0.31 & -0.22 & -7.42 & \textbf{-7.49} \\
& {$\epsilon=4$} & -1.01 & -0.95 & -9.40 & \textbf{-9.74} \\
& {$\epsilon=6$} & -1.85 & -1.86 & -10.54 & \textbf{-10.97} \\
& {$\epsilon=8$} & -3.30 & -3.22 & -12.33 & \textbf{-12.45} \\
\midrule
\multirow{4}{*}{PGD-10} & {$\epsilon=2$} & -0.19 & -0.15 & -36.55 & \textbf{-37.55} \\
& {$\epsilon=4$} & -0.70 & -0.77 & -43.84 & \textbf{-43.93} \\
& {$\epsilon=6$} & -1.88 & -2.26 & -45.31 & \textbf{-45.69} \\ 
& {$\epsilon=8$} & -3.24 & -3.58 & -46.88 & \textbf{-47.08} \\
\bottomrule
\end{tabular}
\caption{Comparison of impact of different task loss-based attacks on model performance (mAP) under various attack sizes using FGSM and PGD-10 on COCO\_traffic. $\mathcal{A}_{loc}$, $\mathcal{A}_{cls}$, $\mathcal{A}_{obj+loc+cls}$, and $\mathcal{A}_{obj}$ are defined similarly as in Table~\ref{tab:kitti att. size}. The clean mAP on COCO\_traffic is 66.10\%.}
\label{tab:coco att. size}
\end{table}

According to the results, we observe that performance degradation due to the adversarial perturbations is different across various task losses, attack types and strengths. Attacks sourced from the objectness loss cause the most performance degradation in both the FGSM and PGD cases. On both datasets, the gaps can be large between the objectness-aware and objectness-unaware cases (e.g., 45\% and 43\% for PGD-10 when $\epsilon = 8$).
In addition, as expected, stronger attacks make the model performance drop more. 

\subsection{Qualitative Analysis of Attacks}
\begin{figure*}
\centering
\begin{tabular}{@{} cccc @{}}
  \toprule
  {Clean} & {$\mathcal{A}_{loc}$(FGSM)} & {$\mathcal{A}_{cls}$(FGSM)} & {$\mathcal{A}_{obj}$(FGSM)} \\
  \midrule
  \multirow{-2.5}{*}{
  \includegraphics[width=.23\linewidth, height=.8in]{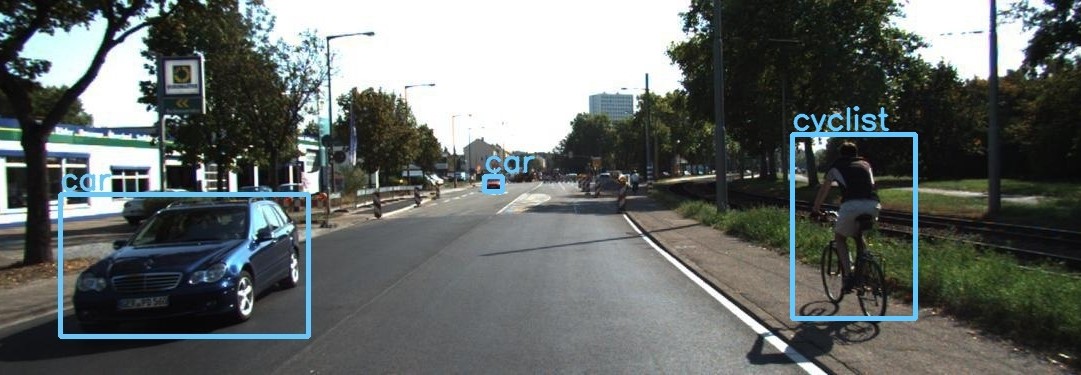}} &
  \includegraphics[width=.23\linewidth, height=.8in]{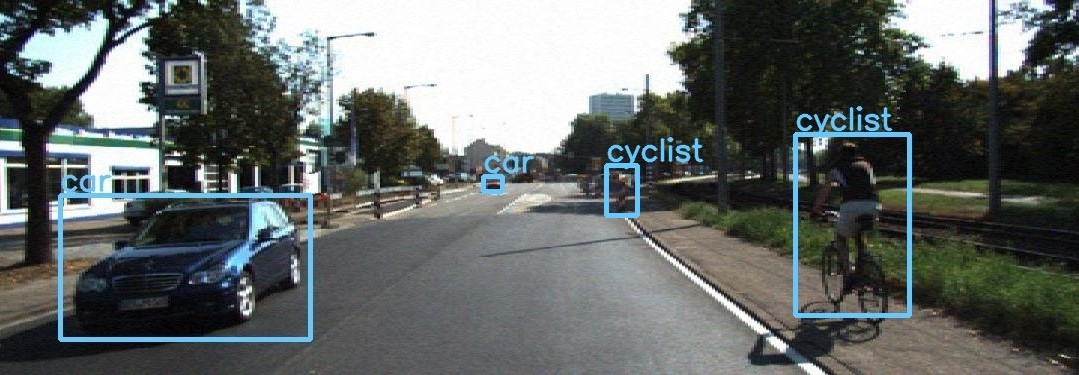} &
  \includegraphics[width=.23\linewidth, height=.8in]{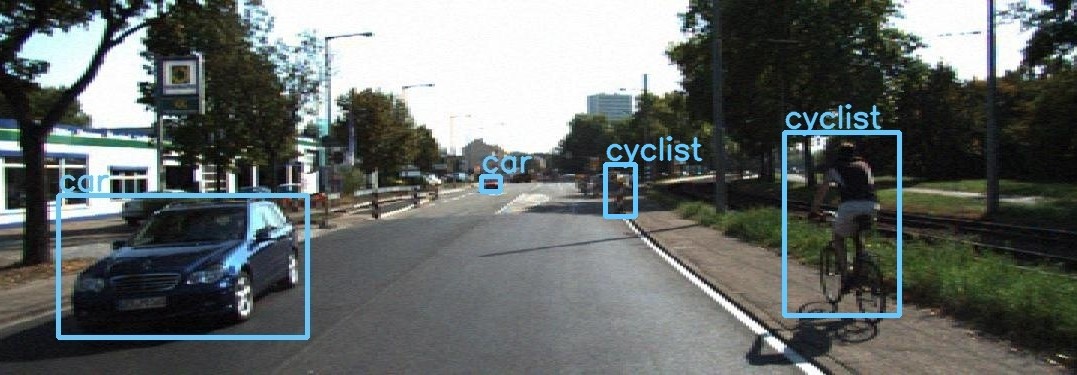} &
  \includegraphics[width=.23\linewidth, height=.8in]{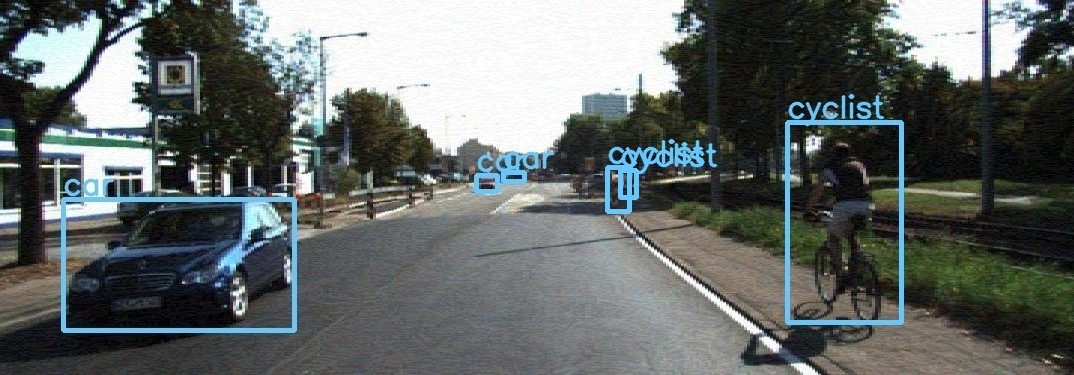} \\
  \cmidrule{2-4}
  & $\mathcal{A}_{loc}$(PGD) & $\mathcal{A}_{cls}$(PGD) & $\mathcal{A}_{obj}$(PGD) \\
  \cmidrule{2-4}
  & \includegraphics[width=.23\linewidth, height=.8in]{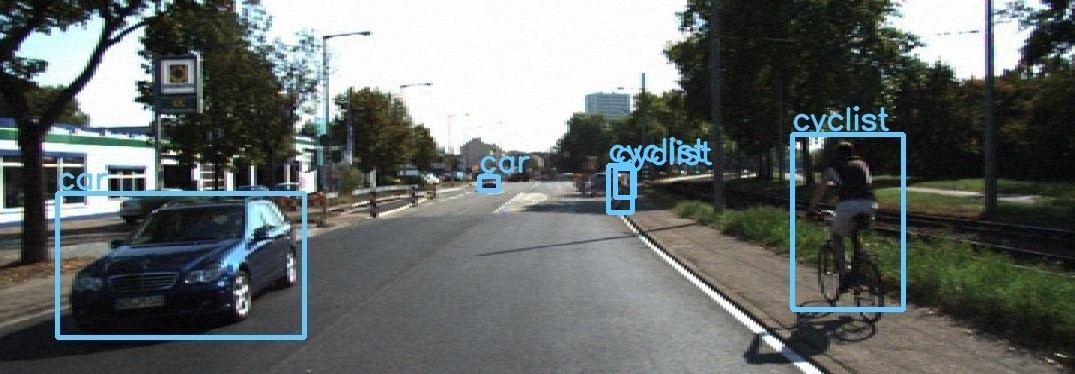}&
  \includegraphics[width=.23\linewidth, height=.8in]{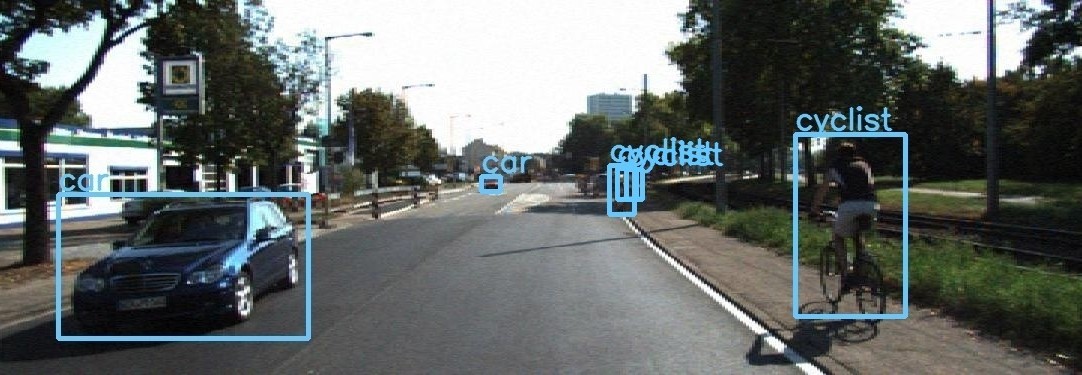}&
  \includegraphics[width=.23\linewidth, height=.8in]{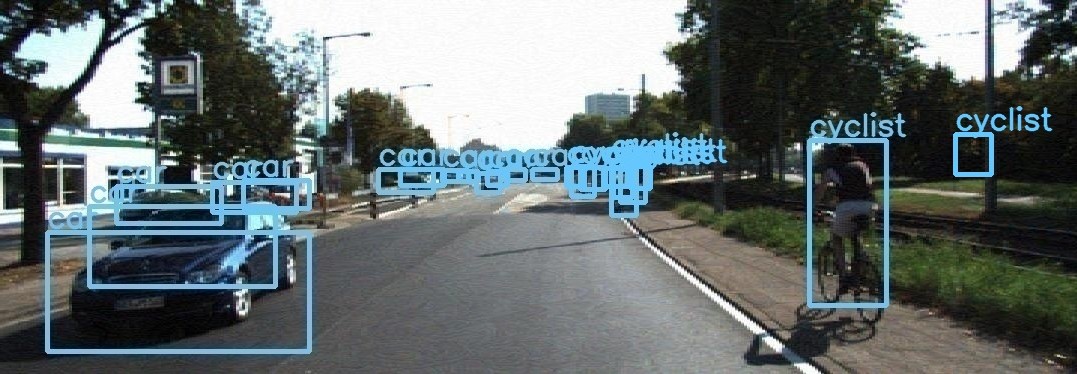}\\
  \bottomrule 
\end{tabular}
\caption{Visualization of detection results under different task-loss-based attacks using FGSM (top row) and 10-step PGD (bottom row) with $\epsilon=4$ on a KITTI example. Best viewed when zoomed in.}
\label{fig:kitti qual. comparison2}
\end{figure*}

\begin{figure*}[t]
\centering
\begin{tabular}{@{} cccc @{}}
  \toprule
  {Clean} & {$\mathcal{A}_{loc}$(FGSM)} & {$\mathcal{A}_{cls}$(FGSM)} & {$\mathcal{A}_{obj}$(FGSM)} \\
  \midrule
  \multirow{-2.5}{*}{
  \includegraphics[width=.23\linewidth, height=.8in]{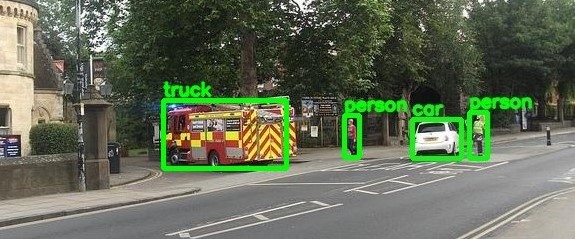}} &
  \includegraphics[width=.23\linewidth, height=.8in]{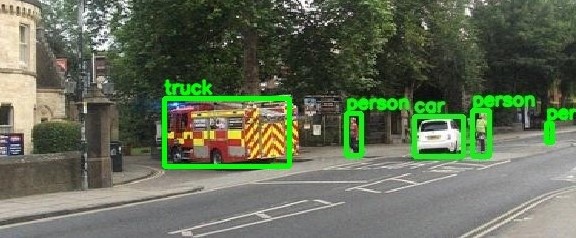} &
  \includegraphics[width=.23\linewidth, height=.8in]{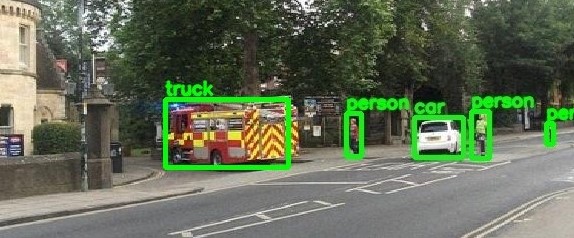} &
  \includegraphics[width=.23\linewidth, height=.8in]{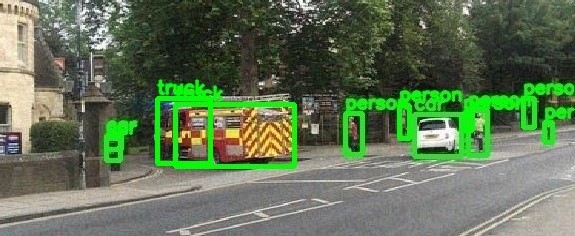} \\
  \cmidrule{2-4}
  & $\mathcal{A}_{loc}$(PGD) & $\mathcal{A}_{cls}$(PGD) & $\mathcal{A}_{obj}$(PGD) \\
  \cmidrule{2-4}
  & \includegraphics[width=.23\linewidth, height=.8in]{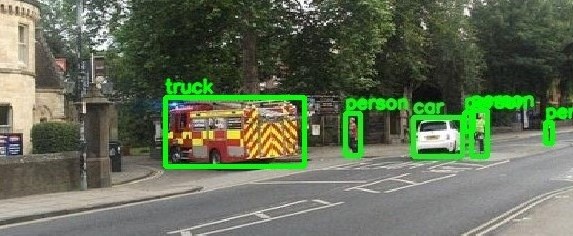}&
  \includegraphics[width=.23\linewidth, height=.8in]{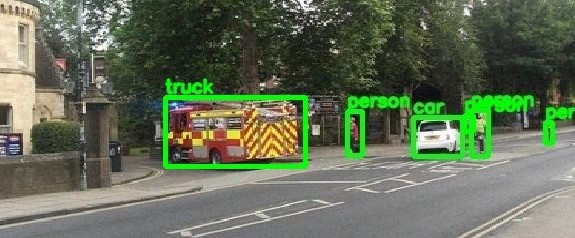}&
  \includegraphics[width=.23\linewidth, height=.8in]{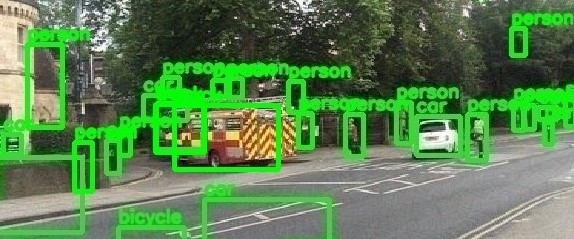}\\
  \bottomrule 
\end{tabular}
\caption{Visualization of detection results under different task-loss-based attacks using FGSM (top row) and 10-step PGD (bottom row) with $\epsilon=4$ on a COCO\_traffic example. Best viewed when zoomed in.}
\label{fig:coco qual. comparison}
\end{figure*}

The qualitative impact of the three attacks (i.e., $\mathcal{A}_{obj}$, $\mathcal{A}_{loc}$, and  $\mathcal{A}_{cls}$) on the model performance are illustrated in Figure~\ref{fig:kitti qual. comparison2} for a KITTI example and Figure~\ref{fig:coco qual. comparison} for a COCO\_traffic example. Comparing with the detection results on the clean image, we observe that many false positives are produced under the objectness-oriented attack on both datasets. 
Sometimes, the attacks from the other two losses can also result in some false positives, but the number is much lower. These qualitative results intuitively demonstrate that the objectness-based attack can be more effective than the other two.
One possible reason is that both classification and localization depend heavily on objectness estimation. Object-oriented attacks can impact both other aspects simultaneously (Figure~\ref{fig:kiiti t-SNE}).
In addition, we can see that our PGD-based objectness attacking strategy is more effective than the FGSM-based one on both datasets, which agrees with what was previously found in the quantitative analysis.

\subsection{Adversarial Training Results}
In this subsection, we further investigate whether a model adversarially trained with the objectness-oriented attacks can be more robust than those trained without. The experiments were also conducted on the KITTI and COCO\_traffic datasets. Particularly, we are interested in the robustness of the following YOLOv4 models trained with attacks derived from different task losses: the model normally trained with only clean images ($\mathcal{M}_{STD}$), the model adversarially trained using the overall loss ($\mathcal{M}_{ALL}$), the model trained with the algorithm in \cite{Zhang2019} ($\mathcal{M}_{MTD}$, where $MTD$ represents multi-task domain), the models trained with adversarial examples solely from one kind of loss ($\mathcal{M}_{LOC}$, $\mathcal{M}_{CLS}$, and $\mathcal{M}_{OBJ}$, where the subscripts respectively represent localization, classification, and objectness), and the model obtained from Algorithm~\ref{alg:Adv.Train} ($\mathcal{M}_{OA}$).
We test these models' robustness under the PGD-based attacks induced from the objectness loss and the overall loss (as these two attacks are shown to be much more destructive in Table~\ref{tab:kitti att. size} and Table~\ref{tab:coco att. size}). The results are reported in Table~\ref{tab:AT_tables}.

As we can see from Table~\ref{tab:AT_tables}, our proposed adversarial defense approaches considering the objectness aspect ($\mathcal{M}_{OBJ}$ and $\mathcal{M}_{OA}$) lead to more robustness than those do not (e.g., $\mathcal{M}_{LOC}$ and $\mathcal{M}_{CLS}$). For example, on KITTI, the mAP of $\mathcal{M}_{OBJ}$ is up to 21\% higher than that of $\mathcal{M}_{STD}$, and on COCO\_traffic, the mAP of $\mathcal{M}_{OA}$ is improved by up to 12.6\% from the baseline under the objectness-oriented PGD attack. 
While we can see that the objectness-awareness plays a critical role in both KITTI and COCO\_traffic cases, the models with the best performance on the two datasets are different (i.e., $\mathcal{M}_{OBJ}$ for KITTI and $\mathcal{M}_{OA}$ for COCO\_traffic). 
One possible reason is that the misalignment of gradients sourced from classification and localization is more serious on KITTI than on COCO\_traffic. It follows that improving one kind of robustness (classification/localization) will be more likely to hurt the other (localization/classification) on KITTI. In this case, we can be better off during adversarial training by ignoring the two disjoint task domains and focusing only on the objectness domain that `overlaps' the other two.

From Table~\ref{tab:AT_tables}, we can also see that our objectness-aware solution ($\mathcal{M}_{OA}$ or $\mathcal{M}_{OBJ}$) outperforms $\mathcal{M}_{MTD}$~\cite{Zhang2019}. The gaps are more obvious on KITTI than on COCO\_traffic. This can also be explained by the hypothesis that the problem of the misalignment of task gradients is less serious on COCO\_traffic than on KITTI.

\begin{table}[t]
\centering
\subfloat[KITTI\label{tab:kitti AT}]{
\centering
\setlength\tabcolsep{3pt}
   \begin{tabular}{@{} l*{2}{S} @{}}
      \toprule
      {Model} & {$\mathcal{A}_{obj}$} & {$\mathcal{A}_{obj+loc+cls}$} \\ 
      \midrule
      $\mathcal{M}_{STD}$ & 28.43 & 28.63 \\ 
      $\mathcal{M}_{ALL}$ & 39.65 & 40.65  \\
      $\mathcal{M}_{MTD}$ & 36.13 & 35.94  \\
      $\mathcal{M}_{LOC}$ & 37.86 & 37.61 \\ 
      $\mathcal{M}_{CLS}$ & 39.29 & 39.70 \\
      $\mathcal{M}_{OBJ}$ & {\textbf{49.43}} & {\textbf{48.83}} \\ 
      $\mathcal{M}_{OA}$ & 42.26 & 41.86  \\
      \bottomrule
   \end{tabular}
} 
\quad
\subfloat[COCO\_traffic\label{tab:coco AT}]{
\centering
\setlength\tabcolsep{3pt}
   \begin{tabular}{@{} l*{2}{S} @{}}
      \toprule
      {Model} & {$\mathcal{A}_{obj}$} & {$\mathcal{A}_{obj+loc+cls}$} \\ 
      \midrule
      $\mathcal{M}_{STD}$ & 22.17 & 22.29 \\ 
      $\mathcal{M}_{ALL}$ & 34.58 & 33.44 \\
      $\mathcal{M}_{MTD}$ & 33.26 & 33.20  \\
      $\mathcal{M}_{LOC}$ & 33.23 & 32.10 \\ 
      $\mathcal{M}_{CLS}$ & 31.71 & 31.58 \\
      $\mathcal{M}_{OBJ}$ & 33.30 & 32.69 \\ 
      $\mathcal{M}_{OA}$ & {\textbf{34.77}} & {\textbf{33.61}}  \\
      \bottomrule
   \end{tabular}
}
\caption{mAP comparison of various adversarially trained YOLO models under PGD-10 attacks on (a) KITTI and (b) COCO\_traffic validation sets. Depending on which losses the adversarial examples are originated from, the following adversarially trained models are obtained for each dataset: $\mathcal{M}_{STD}$, $\mathcal{M}_{ALL}$, $\mathcal{M}_{MTD}$, $\mathcal{M}_{LOC}$, $\mathcal{M}_{CLS}$, $\mathcal{M}_{OBJ}$, $\mathcal{M}_{OA}$ (the notation is explained in the text). $\epsilon=4$.} \label{tab:AT_tables}
\end{table}

\section{CONCLUSION} \label{conclusion}

In this paper, we have identified a serious vulnerability of YOLO detectors in autonomous driving scenarios. The vulnerability comes from the objectness aspect of the object detection. To better understand and to remedy the vulnerability, we have proposed: (1) a new attack strategy targeting the objectness loss in object detection, and (2) an objectness-aware adversarial training framework to enhance the robustness of the detector. 
Additionally, we find that the direction of the image gradient derived from the objectness loss is more consistent with those from the two other losses. Adversarial training considering the objectness aspect can potentially alleviate the problem of misaligned task gradients. Our experiments on the KITTI and COCO\_traffic datasets demonstrate that the objectness-oriented attack approach is much more effective than the attacks derived from the other two detection losses. Furthermore, the proposed adversarial defense approaches explicitly paying attention to the objectness aspect can improve the detector's robustness by large margins on both datasets.

\section*{Acknowledgment}
This research was partially supported by the National Science Foundation (NSF) under Award No. 2153404. This work would not have been possible without the computing resources provided by the Ohio Supercomputer Center.


\bibliography{IEEEabrv,object_detection}

\end{document}